\documentclass[lettersize,journal]{IEEEtran}
\usepackage{amsmath,amsfonts}
\usepackage{algorithmic}
\usepackage{algorithm}
\usepackage{array}
\usepackage[caption=false,font=normalsize,labelfont=sf,textfont=sf]{subfig}
\usepackage{textcomp}
\usepackage{stfloats}
\usepackage{url}
\usepackage{verbatim}
\usepackage{graphicx}
\usepackage{cite}
\usepackage{booktabs}
\usepackage{algorithm}
\usepackage{amsmath}
\usepackage{amssymb}
\usepackage{mathtools}
\usepackage{amsthm}
\usepackage{multirow}

\begin{document}
	
	\newtheorem{define}{Lemma}
	
	\title{Decorrelating Structure via Adapters Makes Ensemble Learning Practical for Semi-supervised Learning}
	
	\author{Jiaqi Wu, Junbiao Pang, Qingming Huang,~\IEEEmembership{Fellow,~IEEE}
		
		\IEEEcompsocitemizethanks
		{
			\IEEEcompsocthanksitem J. Pang, and J.Wu are with the Faculty of Information Technology, Beijing University of Technology, Beijing 100124, China (e-mail: \mbox{junbiao\_pang@bjut.edu.cn}).
			
			\IEEEcompsocthanksitem  Q. Huang is with the University of Chinese Academy of Sciences, Beijing 100049, China.}
		
	}
	
	\maketitle
	
	\begin{abstract}
		In computer vision, traditional ensemble learning methods exhibit either a low training efficiency or the limited performance to enhance the reliability of deep neural networks. In this paper, we propose a lightweight, loss-function-free, and architecture-agnostic ensemble learning by the Decorrelating Structure via Adapters (DSA) for various visual tasks. Concretely, the proposed DSA leverages the structure-diverse adapters to decorrelate multiple prediction heads without any tailed regularization or loss. This allows DSA to be easily extensible to architecture-agnostic networks for a range of computer vision tasks. Importantly, the theoretically analysis shows that the proposed DSA has a lower bias and variance than that of the single head based method (which is adopted by most of the state of art approaches). Consequently, the DSA makes deep networks reliable and robust for the various real-world challenges, \textit{e.g.}, data corruption, and label noises. Extensive experiments combining the proposed method with FreeMatch achieved the accuracy improvements of 5.35\% on CIFAR-10 dataset with 40 labeled data and 0.71\% on CIFAR-100 dataset with 400 labeled data. Besides, combining the proposed method with DualPose achieved the improvements in the Percentage of Correct Keypoints (PCK) by 2.08\% on the Sniffing dataset with 100 data (30 labeled data), 5.2\% on the FLIC dataset with 100 data (including 50 labeled data), and 2.35\% on the LSP dataset with 200 data (100 labeled data).
	\end{abstract}
	
	
	\section{Introduction}
	\label{sec:Introduction}
	
	In the field of computer vision, particularly in Semi-Supervised Learning (SSL), methods such as FixMatch~\cite{sohn2020fixmatch}, FlexMatch~\cite{zhang2021flexmatch}, FreeMatch~\cite{wang2022freematch}, FlatMatch~\cite{huang2023flatmatch}, MaxMatch~\cite{jiang2022maxmatch}, and DualPose~\cite{xie2021empirical} encounter the challenge of the accumulated noisy labels when the pseudo-labeling technique is used in the self-training process. Consequently, SSL should exhibit robustness against the noisy labels. 
	
	In supervised learning tasks, model stability and robustness are crucial to maintain prediction accuracy in the presence of noises, whether from labels or data corruption. Ensemble learning\cite{ren2016ensemble, zhang2017benchmarking, bachman2014learning, wu2024decomposed,  dietterich2000ensemble, brown2004diversity, ju2018relative} emerges as an effective strategy to significantly enhance model robustness. Recently, Channel-based Ensemble (CBE)\cite{wu2024channel}, representing the current State Of The Art (SOTA) in ensemble learning, has been proposed to balance between efficiency and memory cost, demonstrating superior stability and generalization capabilities against input perturbations and data uncertainties.
	
	However, CBE~\cite{wu2024channel} heavily relies on the theoretically guaranteed decorrelation loss function to obtain ensemble diversity. This approach poses several drawbacks as follows:
	\begin{itemize}
		\item \textbf{hyperparameter tuning}: The decorrelation loss function in~\cite{wu2024channel} adds complexity to the hyper-parameter tuning. Improper hyper-parameters not only potentially increase model instability but also adversely affect ensemble capabilities.
		\item \textbf{Over-decorrelation}: When decorrelation loss functions dominate the optimization process, the decorrelation loss would conflict with the task-specific loss. The conflict would damage the plug-in and play ability of CBE.
		\item \textbf{Computational complexity}: The decorrelation loss function naturally introduces additional computational overhead during the model training process. For instance, the computational cost in CBE is linear with respect to the number of prediction heads.
	\end{itemize}
	
	In this paper, motivated by the theoretical analysis of CBE~\cite{wu2024channel}, we introduce a lightweight, loss-function-free, and architecture-agnostic ensemble learning approach, termed the Decorrelation Structure Approach (DSA). The DSA leverages a multi-head prediction structure and incorporates an adapter mechanism~\cite{rebuffi2017learning} into itself. Specifically, the DSA adds an  adapter in front of each prediction head, utilizing these structure-diverse adapters to map the input features into different spaces, thereby reducing the correlation between the prediction heads. As shown in Fig.~\ref{fig:DSA_featuremap}, even with the same loss function, structure-diverse adapters can obtain different mappings. The adapter-based mapping operates on the entire feature, effectively replacing the traditional decorrelation loss function. As a result, DSA offers the following significant advantages as follows:   
	\begin{itemize}
		\item DSA avoids complex hyperparameter tuning processes and mitigates the potential issue of over-decorrelation that may arise from the tuning process.
		\item DSA eliminates potential interference between the decorrelation loss function and the task-specific loss functions, thereby enhancing the stability and generalization ability of the model.
		\item DSA obviates the need for complex correlation calculations, effectively reducing computational costs and improving the training efficiency of the model.
	\end{itemize}
	
	Our method is inherently compatible with both classification and regression tasks and can seamlessly integrate into various SSL frameworks, \textit{e.g.}, FreeMatch, and DualPose. We showcase remarkable performance on pose estimation datasets (\textit{e.g.}, Sniffing, FLIC, LSP) and classification datasets (\textit{e.g.,} CIFAR-10/100). Our primary contributions are as follows:
	
	\begin{itemize}
		\item We propose a structure-based decorrelation method to diminish inter-head correlations for ensemble learning. This method is lightweight, loss-function-free, and architecture-agnostic for deep convolutional neural networks. On noisy dataset, Animal-10N~\cite{song2019selfie}, DSA achieved performance improvement of 1.44\%, over the Multi-Head Ensemble (MHE). These results underscore DSA's ability to suppress the adverse effects of label noise and enable more precise predictions.
		\item The combination of SSL and DSA can effectively improve the accuracy and stability of predictions. In SSL classification tasks, DSA surpassed baseline models, achieving 5.35\% and 0.71\% increases in prediction accuracy on the CIFAR-10 (including 40 labels) and CIFAR-100 (including 400 labels) datasets, respectively. In SSL regression tasks, DSA exhibited outstanding performance, specifically improving prediction accuracy by 2.08\%, 5.2\%, and 2.35\% on the Sniffing (100 data, including 30 labels), FLIC (100 data, including 50 labels), and LSP (200 data, including 100 labels) datasets, respectively, compared to the CBE~\cite{wu2024channel}.
	\end{itemize}
	
	\section{Related Work}
	\label{sec:Related_Work}
	
	\subsection{Learning with Noisy Labels}
	\label{ssec:Noisy_Labels}
	Three prevalent strategies are employed to address noisy labels as follows:
	
	\begin{figure*}[t!]
		\centering
		\subfloat[ME]{\includegraphics[width=1.8in]{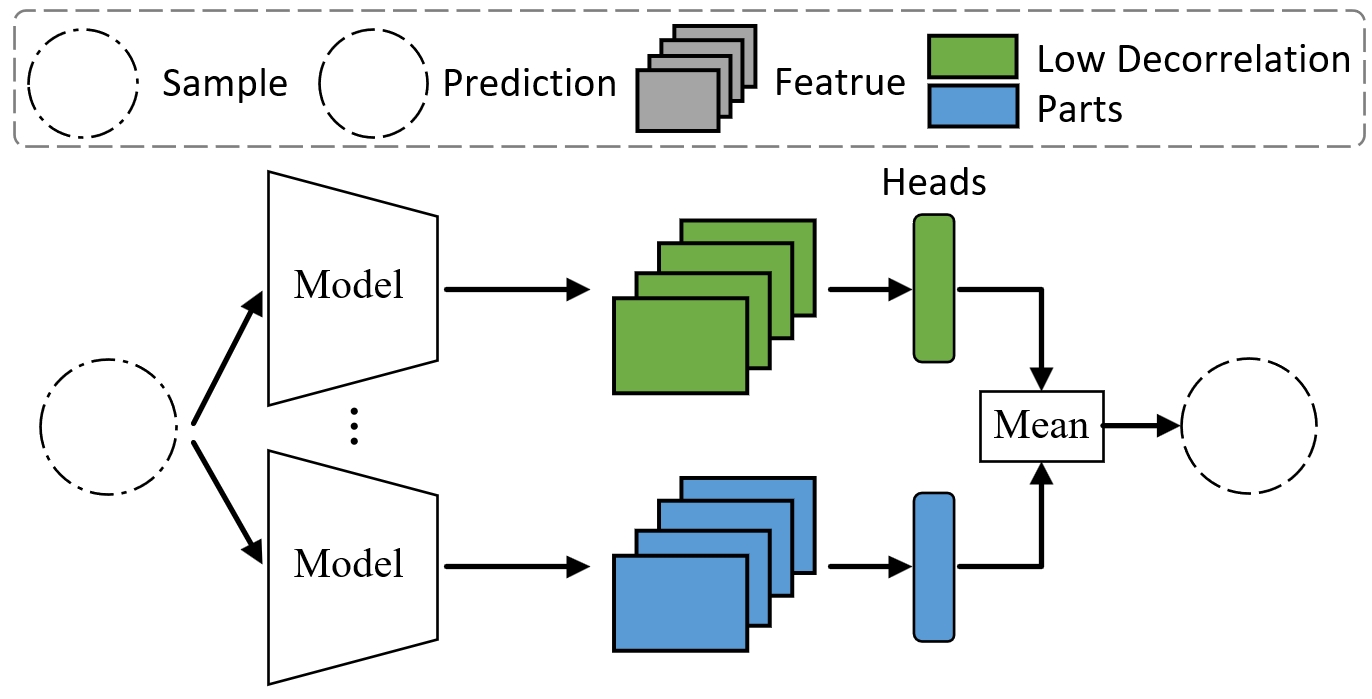} \label{fig:Model_Ensemble_structure}}
		\subfloat[MHE]{\includegraphics[width=1.8in]{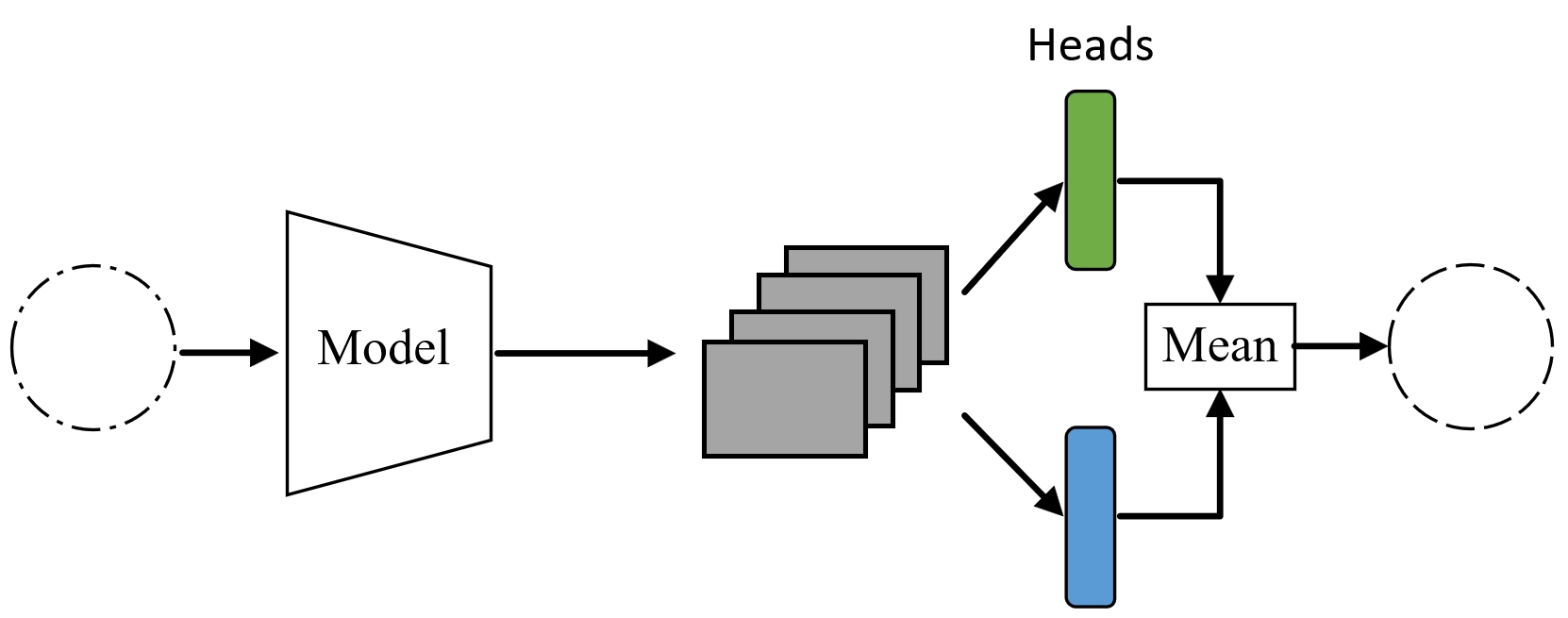} \label{fig:MultiHead_Ensemble_Structure}}
		\subfloat[CBE]{\includegraphics[width=1.8in]{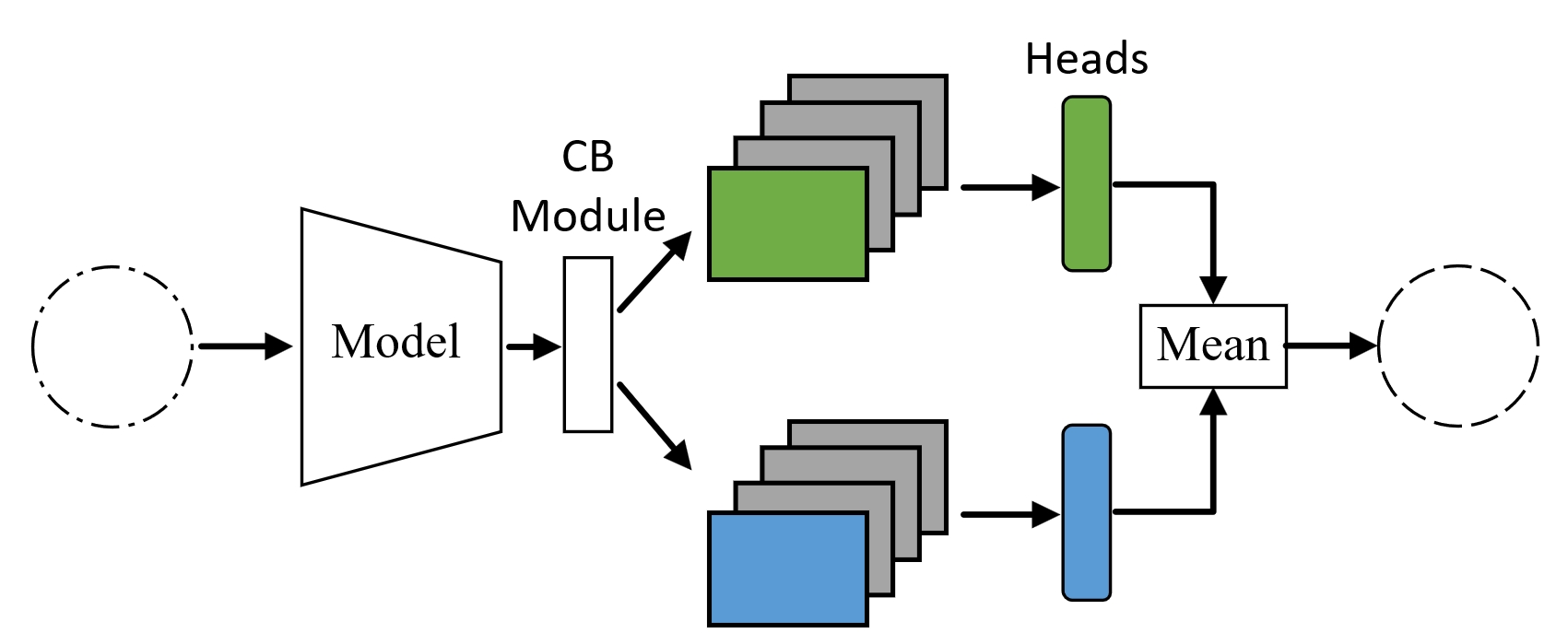} \label{fig:CBE_Structure}}
		\subfloat[DSA]{\includegraphics[width=1.8in]{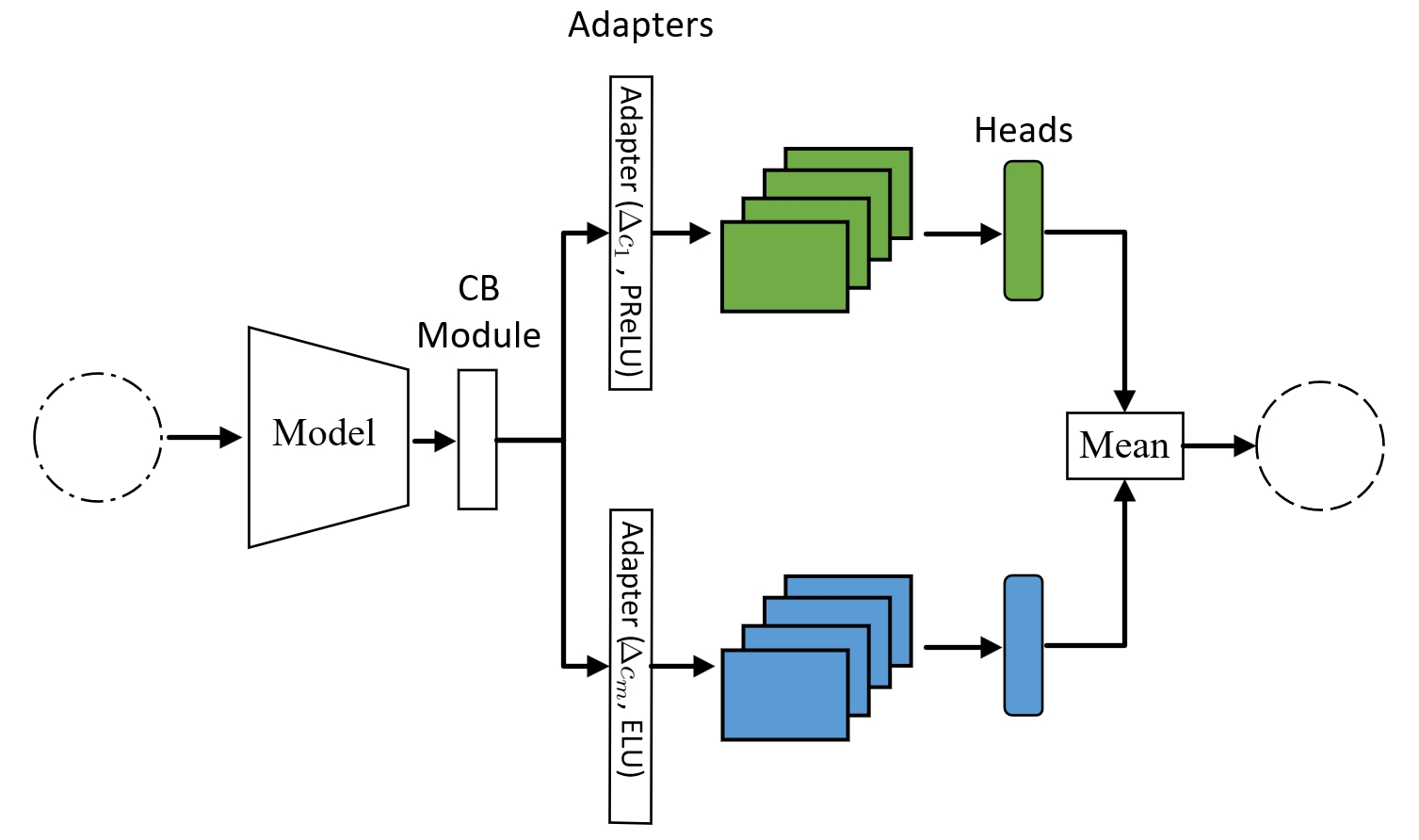} \label{fig:DSA_Structure}}
		\caption{A comparison of structure between the classical ensemble methods and our DSA is presented. The term ``Low Decorrelation Parts'' in the figure denotes the components within each branch that exhibit low correlation due to the influence of the network structure or loss function. In the branch structure, a higher proportion of low-correlation parts contributes to enhancing the accuracy of ensemble predictions.\label{fig:Ensemble_sturcture}}
	\end{figure*}
	
	\textbf{Label correction methods}~\cite{song2019selfie, tanaka2018joint, yi2019probabilistic, zhang2021learning} corrected potential errors by substituting them with more reliable labels, thereby enhancing the quality of dataset labels and improving the performance of machine learning models. One limitation of this method was that the process of replacing noisy labels often required additional data processing costs and time to ensure the acquisition of more accurate labels.
	
	\textbf{Sample re-weighting methods}~\cite{han2018co, jiang2018mentornet, malach2017decoupling} served as an effective strategy for addressing noisy labels. The fundamental concept involved assigning increased importance, or higher weights, to samples with a higher likelihood of having cleaner labels, thereby mitigating the influence of noisy labels during the training phase. A drawback of this method was its heavy reliance on the accuracy of weight assessment. Any inaccuracy in the assessment may potentially have resulted in a decrement of model performance.
	
	\textbf{Overfitting prevention methods}~\cite{ma2018dimensionality, nishi2021augmentation} aimed to prevent networks from fitting too closely to noisy training data, thereby enhancing their ability to generalize to clean test sets. To combat overfitting, a variety of techniques such as regularization, dropout, data augmentation, and early stopping had been proposed. These methods aimed to reduce the model's dependence on noisy data by controlling model complexity, increasing data variety, or introducing randomness during the training process. However, these methods were typically designed for specific tasks, making it difficult to effectively generalize them to other scenarios.
	
	\subsection{Ensemble Learning}
	\label{ssec:Ensemble_Learning}
	
	Existing ensemble learning methods are categorised into three lines: the Model Ensemble (ME), Multi-head Ensemble (MHE), and Channel-based Ensemble (CBE).
	
	\textbf{ME.} Fig.~\ref{fig:Model_Ensemble_structure} illustrates the configuration of Model Ensemble (ME) techniques~\cite{ke2019dual, tang2021humble}. ME employed several complete models to establish an ensemble framework, utilizing combined predictions as the ultimate output. It could be observed that low correlation was achieved in both the prediction heads and their input features. However, the substantial parameter overhead and extensive training durations associated with ME rendered its application unfeasible in practical settings.
	
	\textbf{MHE.} As described in~\cite{wu2024decomposed}, the Multi-Head Ensemble (MHE) method simply implemented ensemble learning by constructing multiple prediction heads, as illustrated in Fig.~\ref{fig:MultiHead_Ensemble_Structure}. Compared to Model Ensemble (ME), MHE offered advantages such as fewer model parameters and shorter training duration. However, in MHE, only the prediction heads achieved low correlation, which led to the issue of homogeneous predictions (as shown in Fig.~\ref{fig:DSA_prediction_similarity}). This phenomenon became particularly evident when the number of iterations exceeded a certain threshold, resulting in a decline in ensemble gain.
	
	\textbf{CBE.} As illustrated in Fig.~\ref{fig:CBE_Structure}, to address the issue of homogeneous predictions, the CBE~\cite{wu2024channel} method built upon the Multi-Head Ensemble (MHE) by splitting and reorganizing features to construct differentiated features that reduced correlation between them. Consequently, CBE achieved low correlation in both the prediction heads and their input features (partial channels).
	
	\textbf{DSA.} As shown in Fig.~\ref{fig:DSA_Structure}, based on CBE, DSA reduced the correlation between shared channels by introducing adapters, achieving decorrelation for complete features. As a result, DSA achieved a comparable level of decorrelation to ME with only half of the model parameters.
	
	\subsection{Decorrelation Methods in Ensemble Learning}
	\label{ssec:Decorrelation_Methods_in_Ensemble_Learning}
	
	\textbf{Random sampling.} Ensemble learning often employed random sampling methods~\cite{breiman1996bagging, ho1998random, dietterich2000ensemble} to diversify models. This approach randomly sampled several training subsets and further trained multiple models. Diversity from different data subsets enhanced the ensemble's robustness. However, random sampling risked losing crucial data, especially in small datasets. Consequently, decorrelation largely depended on the similarity among sampled subsets, which impacted ensemble performance.
	
	\textbf{Heterogeneous model.} Heterogeneous model approaches~\cite{brown2004diversity, zhang2013review, dietterich2000ensemble} involved using structurally diverse models in an ensemble. This diversity included varying architectures, parameters, or training methods, aimed at enhancing ensemble performance by incorporating diverse perspectives. However, this approach often required extensive experimentation to select suitable models, increasing complexity and training costs. Performance variations among models could have led to some models making minimal contributions or even degrading overall performance. Additionally, adding more models could have escalated computational and storage requirements.
	
	\textbf{Adapter.} Adapter methods~\cite{rebuffi2017learning, pfeiffer2020adapterfusion, ruckle2020adapterdrop, karimi2021compacter, wang2020k} were primarily applied in the field of transfer learning. Transfer learning typically involved using a pre-trained model and fine-tuning it for specific tasks. Adapter methods achieved this by inserting adapter modules into the pre-trained model and only updating the parameters of these modules during fine-tuning for downstream tasks, thereby preserving the original parameters of the pre-trained model.
	
	The adapter is a feature mapping mechanism that maps features to the target domain. DSA uses this adapter mechanism for feature mapping. Specifically, an adapter is added before each prediction head, mapping input features into different spaces to reduce correlation between prediction heads. Compared to CBE, which only processes partial input features using a decorrelation loss function, DSA performs comprehensive processing on all input features, achieving a more effective reduction in correlation. 
	
\begin{figure}[t]
		\centering
		\includegraphics[width=3.5in]{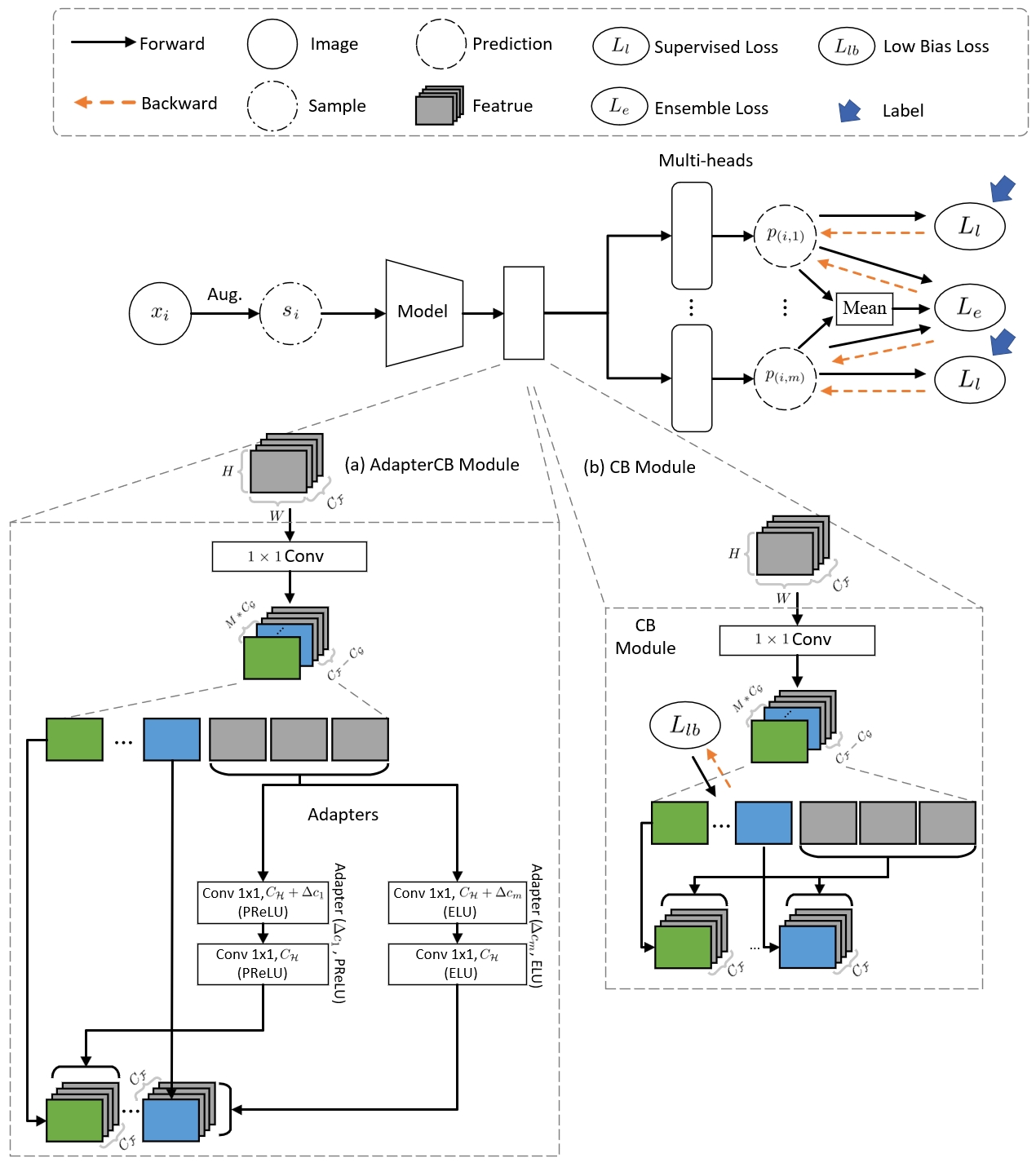}
		\caption{The neural network structure of DSA is as follows. The basic CBE, as depicted in (b), reduces correlation through the use of Low Bias loss. Our proposed AdapterCB module, as depicted in (a), further reduces correlation by incorporating a set of adapters. \label{fig:Network_structure} }
	\end{figure}

	Notably, in DSA, each adapter has a unique structure to avoid degradation, as shown in Fig.~\ref{fig:Network_structure}(b). Specifically, we vary (1) the output channel count of the first convolutional layer in each adapter and (2) the activation function used in each adapter.
	
	\section{Method}
	\label{sec:Method}
	
	\subsection{Revisit CBE}
	
	Fig.~\ref{fig:Network_structure}(b) illustrates that the CBE employs the $1\times1$ convolutional kernel to expand the feature map $\mathcal{F}$ ($\mathcal{F}\in \mathbb{R}^{C_{\mathcal{F}}\times H\times W}$) to $\mathcal{F}_{\text{exp}}$ ($\mathcal{F}_{\text{exp}} \in \mathbb{R}^{(C_{\mathcal{F}}+(M-1)C_{\mathcal{G}})\times H\times W}$). $\mathcal{F}_{\text{exp}}$ can be divided into two parts, one shared feature map $\mathcal{H}$ ($\mathcal{H} \in \mathbb{R}^{C_{\mathcal{H}} \times H\times W}$, where $C_{\mathcal{H}} = C_{\mathcal{F}}-C_{\mathcal{G}}$) and $M$ private feature maps $\{\mathcal{G}_m\}^{M}_{m=1}$ ($\mathcal{G}_m \in \mathbb{R}^{C_{\mathcal{G}}\times H\times W}$). 
	
	Then, the $M$ output feature maps $\{\mathcal{F}^{LD}_m\}^{M}_{m=1}$, which are generated by concatenating the shared feature map with each of the $M$ private feature maps, are input into $M$ prediction heads. The $m$-th output feature map of $\{\mathcal{F}^{LD}_m\}^{M}_{m=1}$ can be expressed as:
	
	\begin{equation}\label{eq:IndependentOutputFeature_CBE}
		\mathcal{F}^{LD}_m= \text{concat}(\mathcal{H}, \mathcal{G}_m) ,
	\end{equation}
	where $\text{concat}(A, B)$ represents the concatenation of feature maps $A$ and $B$ along the channel dimension.
	
	In the CBE, to maximize the diversity of the ensemble, the Low Bias (LB) loss $L_{lb}$ is employed. The primary role of this loss function is to minimize the correlation among the M private feature maps. Specifically, $L_{lb}$ is designed to achieve this by minimizing the correlation between each head, as detailed below:
	
	\begin{equation}\label{eq:FULoss}
		L_{lb}=\frac{1}{\mu N_B}\sum_{i=1}^{\mu N_B}\frac{1}{M}\sum_{i=1}^{M}\sum_{j=1,j\neq i}^{M}\text{cov}(\mathcal{G}_{i},\mathcal{G}_{j}),
	\end{equation}
	where $ \text{cov}(\mathcal{G}_{i},\mathcal{G}_{j}) $ is the correlation function between the features of two prediction heads, where $\mathcal{G}_{i}$, $\mathcal{G}_{j}$ represent the private features of these two heads.
	
	However, it is important to note that the decorrelation process based on the loss function in CBE only reduces the correlation within the private channel portion of the input features, and does not comprehensively cover all input feature maps. 
	
	Furthermore, due to the influence of the decorrelation loss function, CBE is susceptible to complex hyperparameter tuning processes, risks of over-decorrelation, conflicts in loss function optimization, and significant increases in computational costs.
	
	\subsection{Structure-based Decorrelation}
	\label{ssec:Structure_based_Decorrelation}
	
	In traditional ensemble learning methods, a commonly used strategy to enhance the diversity of the ensemble framework is to combine base models with multiple prediction heads. Inspired by this, the present study introduces a decorrelation mechanism named ``Structure-based Decorrelation only with Adapters (SDoAs)'' that is based on the multi-head ensemble framework.
	
	\textbf{SDoAs.} The structure is shown in Fig.~\ref{fig:Network_structure}(a). We introduce a group of adapters to generate $M$ independent feature maps, denoted as $\{\mathcal{H} + \Delta\mathcal{H}_m\}^{M}_{m=1}$, from the shared feature map $\mathcal{H}$. Here, $\{\Delta\mathcal{H}_m\}^{M}_{m=1}$ represent the output of the $M$ adapters. As shown in Fig.~\ref{fig:Network_structure}(b), to avoid the degradation of the adapter mechanism, each adapter has a unique structure. Specifically, this includes using different numbers of channels in the $1 \times 1$ convolutional layer and employing diverse activation functions. The $m$-th output feature map in the set $ \{\mathcal{F}^{SDoAs}_m\} ^{M}_{m=1} $ can be expressed as:
	\begin{equation}\label{eq:IndependentOutputFeature_SD_noPF}
		\mathcal{F}^{SDoAs}_m= \mathcal{H} + \Delta\mathcal{H}_m, 
	\end{equation}
	where $\Delta\mathcal{H}_m$ represents the output of the $m$-th adapter.
	
	\textbf{Overall decorrelation}: The adapters in SDoAs achieve overall decorrelation for the input features of the prediction heads. Specifically, the adapters add a set of distinct values, $\{\Delta\mathcal{H}_m\}^{M}_{m=1}$, to all channels of the M input features, thereby mapping them into different spaces.
	
	\textbf{DSA.} Given the advantages of CBE in lightweight construction of ensemble frameworks, we combine SDoAs with CBE and name this structure the Decorrelation Structure via Adapters (DSA). In the process of integrating SDoAs with CBE, we retain the original split and reorganization processing strategy for input feature channel dimensions, which is based on the private-shared mechanism in CBE.
	
	In light of the stronger decorrelation potential of SDoAs, we omit the decorrelation loss function for the private channel part in CBE. The $m$-th output feature map of $\{\mathcal{F}^{DSA}_m\}^{M}_{m=1}$ can be expressed as:
	\begin{equation}\label{eq:IndependentOutputFeature_ACBE}
		\mathcal{F}^{DSA}_m= \text{concat}(\mathcal{H} + \Delta\mathcal{H}_m, \mathcal{G}_m)
	\end{equation}
	
	\textbf{Lower computational cost}: In DSA, the decorrelation mechanism consists of two components: $\{\Delta\mathcal{H}_m\}^{M}_{m=1}$ and $\{\mathcal{G}_m\}^{M}_{m=1}$. $\{\Delta\mathcal{H}_m\}^{M}_{m=1}$ are generated by a set of adapters, each comprising two $1 \times 1$ convolutional layers and two activation functions, as depicted in Fig.~\ref{fig:Network_structure}(a). The first layer expands the channel count, while the second restores it to the original number. This design facilitates effective feature mapping. Consequently, each adapter possesses few parameters, resulting in no significant increase in computational cost. To ensure unique mapping effects, we vary (1) the output channel count of the first layer, e.g., the output channel count of the first layer in the $m$-th adapter is $ C_{\mathcal{H}}+\Delta c_m $, where $C_{\mathcal{H}}$ is the channel count of the shared feature, and $\Delta c_m=10*m$ is the additional number, which differs for each adapter, and (2) the activation function used in each adapter. As illustrated in Fig.~\ref{fig:Network_structure}(a), the first adapter uses PReLU, and the last adapter uses ELU.
	
	When DSA employs the set $\{\mathcal{G}_m\}^{M}_{m=1}$, it obviates the need for the decorrelation loss of CBE, thus avoiding complex correlation calculations. The calculation process of the correlation for the LB loss is analogous to Eq.~\eqref{eq:CovarFunction}. Therefore, in an iteration, the time complexity of the LB loss is $O(\mu N_B \times (C_{\mathcal{F}}\times H\times W)^2)$. 
	
	As evident from Table~\ref{tab:Classification_Cost}, FreeMatch+DSA achieves significant improvements over FreeMatch+CBE and FreeMatch+DSA without a substantial increase in computational cost. Furthermore, compared to FreeMatch+CBE, FreeMatch+DSA reduces the Average Training Time Per Epoch (ATTPE) by 1.72 seconds.
	
	\begin{table}[h]
		\caption{The comparison of the computational costs among FreeMatch+MHE, FreeMatch+CBE, and FreeMatch+DSA on CIFAR10@40. The ATTPE was measured on a GPU 2080Ti. \label{tab:Classification_Cost}}
		\centering
		\resizebox{\columnwidth}{!}{%
			\begin{tabular}{@{}lccccc@{}}
				\toprule 
				Method & Model Parameters $\downarrow$ & FLOPs $\downarrow$ & MACs (Millions) $\downarrow$ & ATTPE (Seconds) $\downarrow$ & Error Rate (\%) $\downarrow$ \\
				\midrule
				FreeMatch+MHE & \pmb{1.469}M & \pmb{216.288}M & \pmb{4501} & \pmb{226.17} & 16.51 \\   
				FreeMatch+CBE & 1.473M & 216.390M & 4513 & 231.35 & \underline{11.45} \\ 
				FreeMatch+DSA & 1.473M & 216.390M & 4513 & \underline{229.63} & \pmb{6.10} \\ 
				\bottomrule
			\end{tabular}%
		}
	\end{table}
	
	\subsection{Theoretical Comparison of Decorrelation Methods}
	\label{ssec:Theoretical_Comparison_of_Decorrelation_Methods}
	
	In ensemble learning, ensemble diversity is a crucial factor affecting ensemble gain. To enhance the performance of ensemble prediction, we aim to reduce the correlation between individual predictors within the ensemble framework. Therefore, one natural question arises: Can we theoretically compare the correlations among prediction heads for the three methods (\textit{i.e.}, CBE, SDoAs, and DSA) in Section~\ref{ssec:Structure_based_Decorrelation}?
	
	Before comparing, let's first introduce our method for quantifying correlation. 
	
	Let $\mathcal{F}_1$ ($\mathcal{F}_1 \in \mathbb{R}^{C_{\mathcal{F}}\times H\times W}$) and $\mathcal{F}_2$ ($\mathcal{F}_2 \in \mathbb{R}^{C_{\mathcal{F}}\times H\times W}$) be two features, where $H$, $W$, and $C_{\mathcal{F}}$ denote the height, width, and channel, respectively. When calculating the correlation between $\mathcal{F}_1$ and $\mathcal{F}_2$, we resize these two features to one-dimensional features $\mathcal{F}_1^{'}$ ($\mathcal{F}_1^{'} \in \mathbb{R}^{C_{\mathcal{F}}\times H\times W}$) and $\mathcal{F}_2^{'}$ ($\mathcal{F}_2^{'} \in \mathbb{R}^{C_{\mathcal{F}}\times H\times W}$), and calculate the correlation coefficient of $\mathcal{F}_1^{'}$, $\mathcal{F}_2^{'}$. Thus the correlation between $\mathcal{F}_1$ and $\mathcal{F}_2$ can be expressed as follows:
	
	\begin{equation}\label{eq:CovarFunction}
		\text{cov}(\mathcal{F}_1, \mathcal{F}_2) = \text{corrcoef}(\mathcal{F}_1^{'}, \mathcal{F}_2^{'}) ,
	\end{equation}
	where $ \text{corrcoef}(A, B) $ is used to calculate the correlation coefficient between the two one-dimensional features $A$ and $B$.
	
	To simplify the calculation of correlation for the three methods, we set the number of prediction heads to two. Thus, the correlation $\mathcal{C}_{\mathcal{F}^{LD}}$ of $\{\mathcal{F}^{LD}_m\}^ {M}_{m=1}$ can be denoted as:
	\begin{equation}\label{eq:Covar_LD}
		\begin{aligned}
			\mathcal{C}_{\mathcal{F}^{LD}} = \text{cov}(\text{concat}(\mathcal{H}, \mathcal{G}_1), \text{concat}(\mathcal{H}, \mathcal{G}_2)) ,
		\end{aligned}
	\end{equation}
	where $ \text{cov}(\cdot) $ denotes the covariance function in Eq.~\eqref{eq:CovarFunction}.
	
	Similarly, the correlation $\mathcal{C}_{\mathcal{F}^{SDoAs}}$ of $\{\mathcal{F}^{SDoAs}_m\}^{M}_{m=1}$ can be denoted as:
	\begin{equation}\label{eq:Covar_SDoA}
		\begin{aligned}
			\mathcal{C}_{\mathcal{F}^{SDoAs}}=\text{cov}(\text{concat}(\mathcal{H} + \Delta\mathcal{H}_1), \text{concat}(\mathcal{H} + \Delta\mathcal{H}_2))
		\end{aligned}
	\end{equation}
	
	The correlation $\mathcal{C}_{\mathcal{F}^{DSA}}$ of $\{ \mathcal{F}^ {DSA}_m\}^ {M}_{m=1}$ can be denoted as:
	\begin{equation}\label{eq:Covar_SD}
		\begin{aligned}
			\mathcal{C}_{\mathcal{F}^{DSA}}=\text{cov}(\text{concat}(\mathcal{H} + \Delta\mathcal{H}_1, \mathcal{G}_1), \text{concat}(\mathcal{H} + \Delta\mathcal{H}_2, \mathcal{G}_2))
		\end{aligned}
	\end{equation}
	
	\begin{define}[DSA exhibits lower correlation than CBE.] \label{def:Cov_of_LD}
		Due to $\Delta\mathcal{H}_1 \ll \mathcal{H}$ and $\Delta\mathcal{H}_2 \ll \mathcal{H}$, we approximate $\tilde{H} \approx \mathcal{H} + \Delta\mathcal{H}_1 \approx \mathcal{H} + \Delta\mathcal{H}_2$. The relationship between $\mathcal{C}_{\mathcal{F}^{DSA}}$ and $\mathcal{C}_{\mathcal{F}^{SDoAs}}$ is modeled as follows:
		\begin{equation}\label{eq:Cov_of_LD}
			\begin{aligned}
				\mathcal{C}_{\mathcal{F}^{DSA}} &=\text{cov}(\text{concat}(\mathcal{H} + \Delta\mathcal{H}_1, \mathcal{G}_1), \text{concat}(\mathcal{H} + \Delta\mathcal{H}_2, \mathcal{G}_2)) \\
				& \approx \mathbb{E}\big((\tilde{H}+\mathcal{G}_1)(\tilde{H}+\mathcal{G}_2)\big) - \mathbb{E}(\tilde{H}+\mathcal{G}_1)\mathbb{E}(\tilde{H}+\mathcal{G}_2) \\
				& \approx \text{var}(\tilde{H}) + \text{cov}(\mathcal{G}_1, \mathcal{G}_2) \\
				& \le \text{var}(\mathcal{H}) + \text{cov}(\mathcal{G}_1, \mathcal{G}_2) \\
				& \le \mathcal{C}_{\mathcal{F}^{LD}},
			\end{aligned}
		\end{equation}
		where $\mathbb{E}(\cdot)$ and $ \text{var}(\cdot) $ denote the expectation and variance functions, respectively.
	\end{define}
	
	\textbf{Remarks of Lemma~\ref{def:Cov_of_LD}.} Eq.~\eqref{eq:Cov_of_LD} reveals that the DSA method exhibits lower theoretical correlation compared to LD, which only utilizes the private feature maps.
	
	\begin{define}[DSA exhibits lower correlation than SDoAs.] \label{def:Cov_of_SDoA}
		Since $\Delta\mathcal{H}_1$ and $\Delta\mathcal{H}_2$ are both independent of $\mathcal{G}_1$ and $\mathcal{G}_2$, it follows that $\text{cov}(\Delta\mathcal{H}_1 + \mathcal{G}_1, \Delta\mathcal{H}_2 + \mathcal{G}_2) \le \text{cov}(\Delta\mathcal{H}_1, \Delta\mathcal{H}_2)$. The relationship between $\mathcal{C}_{\mathcal{F}^{DSA}}$ and $\mathcal{C}_{\mathcal{F}^{SDoAs}}$ is modeled as follows:
		\begin{equation}\label{eq:Cov_of_SDoA}
			\begin{aligned}
				\mathcal{C}_{\mathcal{F}^{DSA}} &=\text{cov}(\text{concat}(\mathcal{H} + \Delta\mathcal{H}_1, \mathcal{G}_1), \text{concat}(\mathcal{H} + \Delta\mathcal{H}_2, \mathcal{G}_2)) \\
				& \approx 
				\begin{split}
					\mathbb{E}\big((\mathcal{H}+(\Delta\mathcal{H}_1 + \mathcal{G}_1))(\mathcal{H}+(\Delta\mathcal{H}_2 + \mathcal{G}_2)\big) - \\ \mathbb{E}(\mathcal{H}+(\Delta\mathcal{H}_1 + \mathcal{G}_1))\mathbb{E}(\mathcal{H}+(\Delta\mathcal{H}_2 + \mathcal{G}_2)) \\
				\end{split} \\
				& \approx \text{var}(\mathcal{H}) + \text{cov}(\Delta\mathcal{H}_1 + \mathcal{G}_1, \Delta\mathcal{H}_2 + \mathcal{G}_2) \\
				& \le \text{var}(\mathcal{H}) + \text{cov}(\Delta\mathcal{H}_1, \Delta\mathcal{H}_2) \\
				& \le \mathcal{C}_{\mathcal{F}^{SDoAs}},
			\end{aligned}
		\end{equation}
	\end{define}
	
	\textbf{Remarks of Lemma~\ref{def:Cov_of_SDoA}.} The inequality in Eq.~\eqref{eq:Cov_of_SDoA} indicates that the DSA method exhibits lower theoretical correlation compared to using adapters alone.
	
	\subsection{Application to SSL}
	\label{ssec:Application_to_SSL}
	
	The DSA network is shown in Fig.~\ref{fig:Network_structure}(a). First, given the data $x_i$ and the data augmentation function $\omega(\cdot)$, we generate a sample $s_i$ from $x_i$ using $\omega(\cdot)$. Second, for a sample $x_i$, the DSA generates predictions $\mathcal{P}_i=\{p_{(i,m)}\}^M_{m=1}$ using the $M$ heads.
	
	For the labeled data $\{(x_i,y_i)\}^{N_L}_{i=1}$, the supervised training is guided by the ground truth $y_i$ as follows:
	\begin{small}
		\begin{equation}\label{eq:SupevisedLoss}
			L_{l}=\frac{1}{N_B}\sum_{i=1}^{N_B}\frac{1}{M}\sum_{m=1}^{M}\text{CE}\big(p_{(i,m)}, y_{i}\big) ,
		\end{equation}
	\end{small}
	where $N_B$ denotes the batch size of the labeled data, and $\text{CE}(\cdot,\cdot)$ represents the cross-entropy function. 
	
	For unlabeled data $\{x_i\}^{N_L+N_U}_{i=N_L+1}$, the threshold $\tau$ is applied to filter the unreliable predictions from the prediction scores $\mathcal{P}_i$. The ensemble prediction $\overline{\mathcal{P}}_i$ is calculated as follows:
	\begin{equation}\label{eq:PseudoLabel}
		\overline{\mathcal{P}}_i=\frac{1}{M}\sum_{m=1}^{M}\Gamma(\text{max}(p_{(i, m)}) >\tau)\cdot p_{(i, m)},
	\end{equation}
	where $\Gamma(\cdot>\tau)$ denotes the indicator function for the confidence threshold, where $\tau$ is the specified threshold.
	
	$\overline{\mathcal{P}}_i$ in Eq.~\eqref{eq:PseudoLabel} is further utilized as a pseudo-label to supervise $\mathcal{P}_i$. Hence, the ensemble supervised loss $L_{e}$ is formulated as follows:
	\begin{equation}\label{eq:EnsembleLoss}
		L_{e}=\frac{1}{\mu N_B}\sum_{i=1}^{\mu N_B}\frac{1}{M}\sum_{m=1}^{M}\text{CE}\big(p_{(i,m)},\overline{\mathcal{P}}_i\big),
	\end{equation}
	where $\mu$ represents the ratio of the number of unlabeled data points to the number of labeled data points.
	
	\section{Experiments on SSL Pose Estimation}
	\label{sec:Experiments_on_Pose_Estimation}
	
	SSL 2D pose estimation is utilized to validate the effectiveness of our method on the regression task.
	
	\subsection{Dataset and Evaluation Metrics}
	\label{ssec:Pose_Estimation_Dataset}
	
	\begin{figure}[h]
		\centering
		\subfloat{
			\includegraphics[width=1.1in]{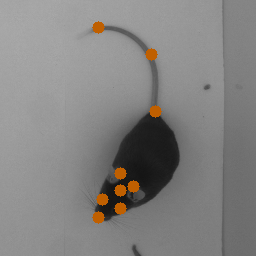}
			\includegraphics[width=1.1in]{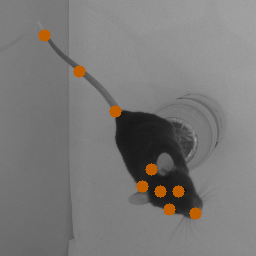}
			\includegraphics[width=1.1in]{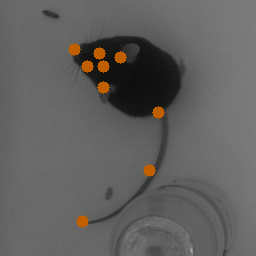}
		}
		\hfil
		\subfloat{
			\includegraphics[width=1.1in]{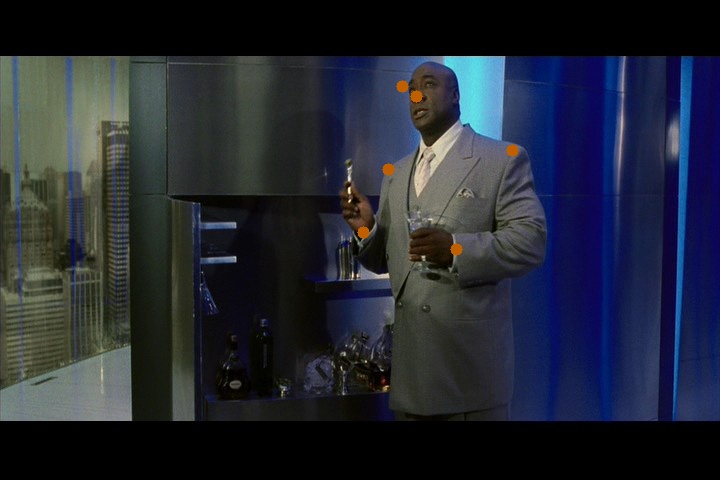}
			\includegraphics[width=1.1in]{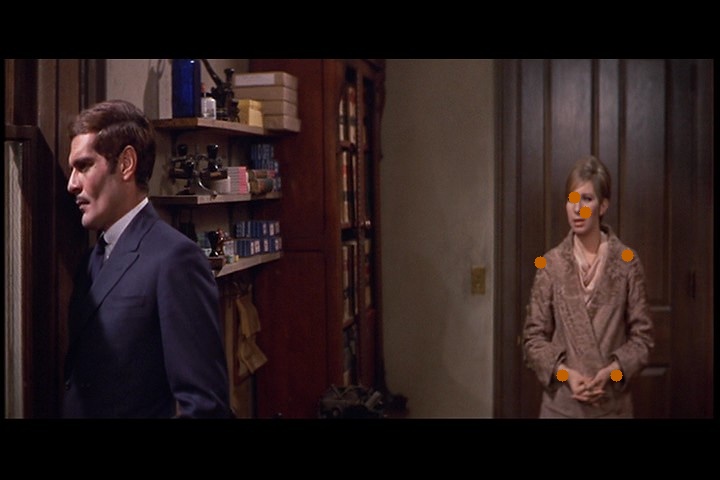}
			\includegraphics[width=1.1in]{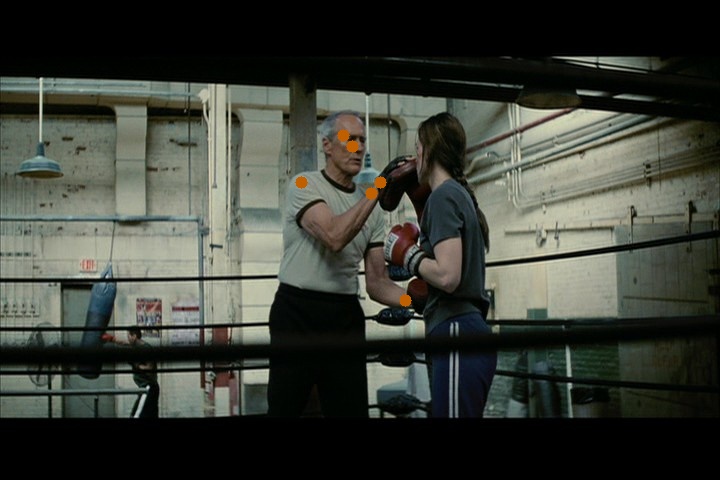}
		}
		\hfil
		\subfloat{
			\includegraphics[width=1.1in]{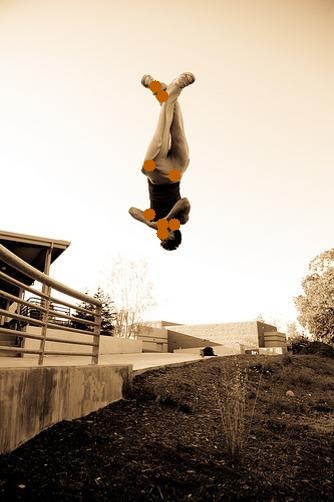}
			\includegraphics[width=1.1in]{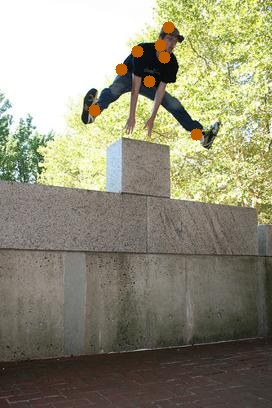}
			\includegraphics[width=1.1in]{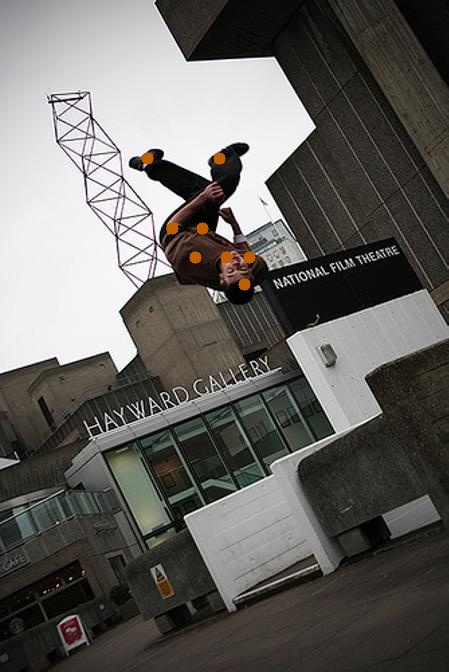}
		}
		\caption{Samples from the Sniffing (top row), FLIC (middle row), and LSP (bottom row) datasets are presented. \label{fig:D_Sample}}
	\end{figure}
	
	Our experiments are primarily conducted on three publicly available datasets: FLIC~\cite{sapp2013modec}, LSP~\cite{johnson2011learning}, and the Sniffing dataset\footnote{\url{https://github.com/Qi2019KB/Sniffing-Dataset}}. The image data and their corresponding labels from the Sniffing dataset are illustrated in Fig.~\ref{fig:D_Sample}.
	
	To precisely evaluate the performance of DSA, we report the metrics of Mean Squared Error (MSE) and Percentage of Correct Keypoints (PCK). Specifically, the mean MSE is defined as follows:
	
	\begin{equation}\label{eq:MSE_formula}
		MSE =  \frac{1}{I} \frac{1}{K} \sum_{i=1}^{I} \sum_{k=1}^{K} \sqrt{|{pred}_{i,k} - gt_{i, k}|^{2}} ,
	\end{equation}
	where $I$, $K$ denote the number of images and keypoints, respectively. $i$, $k$ are the indices of images and keypoints, respectively. ${pred}_{i,k}$ and $gt_{i,k}$ denote the predicted and ground-truth locations of the $k$-th keypoint in the $i$-th image. 
	
	Mean Squared Error (MSE) serves as an intuitive metric for the Euclidean distance between predicted and ground-truth points. A lower MSE value indicates a smaller overall error. The mean of the Percentage of Correct Keypoints (PCK) is defined as follows:
	
	\begin{equation}\label{eq:PCK_formula}
		PCK = \frac{\sum_{i=1}^{I} \sum_{k=1}^{K} \sigma \big (\frac{d_{i,k}}{l} \leq T \big)}{\sum_{i=1}^{I} \sum_{k=1}^{K} 1} ,
	\end{equation}
	where $l$ denotes a normalized distance defined by the protocols of the different datasets. $T$ is a threshold, where $0\leq T \leq 1$, and the function $\sigma(A )$ is defined such that $\sigma(A)=1$ if the condition $A$ is true, and $\sigma(A)=0$ otherwise.
	
	In contrast to MSE, PCK incorporates the variability in label errors across different datasets when assessing the accuracy of pose estimation predictions. A higher PCK value indicates greater accuracy of a method. PCK@$T$ denotes the use of a normalized distance $l$ to scale the MSE and the application of threshold $T$ to determine prediction accuracy. Specifically, for the Sniffing dataset, $l$ represents the distance between the left and right eyes; for the FLIC dataset, $l$ is the distance between the left and right shoulders; and for the LSP dataset, $l$ is the distance between the left shoulder and left hip.
	
	We employ various PCK evaluation criteria on three datasets, including PCK@0.1 ($T=0.1$), PCK@0.2 ($T=0.2$), PCK@0.3 ($T=0.3$), and PCK@0.5 ($T=0.5$), to assess the accuracy of keypoint predictions at different evaluation thresholds.
	
	\subsection{Configurations}
	\label{ssec:Pose_Estimation_Configurations}
	
	In all experiments, the pose model was Stacked Hourglass with the same configuration as in~\cite{newell2016stacked}. All experiments were conducted using the Adam optimizer with a weight decay of 0.999. The learning rate was set to 0.00025. The batch size was 4, with 2 labeled data points.
	
	We conducted SSL pose estimation experiments using DualPose as the SSL method. DualPose employs two data augmentation strategies to generate a set of sample pairs with varying prediction difficulty. The strong data augmentation included: random rotation ($+/- 30$ degrees), random scaling ($ 0.75 - 1.25 $), and random horizontal flip. The weak data augmentation included: random rotation ($+/- 5$ degrees), random scaling ($ 0.95 - 1.05 $), and random horizontal flip.
	
	\subsection{Comparisons}
	\label{ssec:Pose_Estimation_Comparisons}
	
	\begin{table}[h]
		\caption{The MSE and PCK of SSL pose estimation with 200 training epochs are reported. ``Supervised'' denotes a pose model trained with only labeled data. ``X+ CBE'' denotes the CBE network. ``X+ DSA'' denotes our DSA network. ``X/Y'' denotes that X labeled data and Y unlabeled ones are used in SSL, respectively. \label{tab:Pose_Main_Results}}
		\centering
		\resizebox{\columnwidth}{!}{%
			\begin{tabular}{@{}lcccc@{}}
				\toprule 
				\multirow{3}{*}{Method} & \multicolumn{4}{c}{Sniffing} \\
				& \multicolumn{2}{c}{30/100} & \multicolumn{2}{c}{60/200} \\
				& MSE $\downarrow$ & PCK@0.2 (\%) $\uparrow$ & MSE $\downarrow$ & PCK@0.2 (\%)  $\uparrow$\\
				\midrule
				Supervised   & 4.72 $\pm$ 0.15              & 15.28 $\pm$ 2.52             & 4.60 $\pm$ 0.32              & 29.17 $\pm$ 1.73             \\
				\midrule
				\midrule
				DualPose~\cite{xie2021empirical} & 4.96 $\pm$ 0.09 & 16.67 $\pm$ 1.81      & 4.38 $\pm$ 0.07              & 30.56 $\pm$ 1.21             \\
				DualPose+CBE & \pmb{3.59} $\pm$ 0.06        & \underline{67.36} $\pm$ 0.47 & \underline{4.15} $\pm$ 0.07  & \underline{70.34} $\pm$ 0.35 \\
				DualPose+DSA & \underline{3.65} $\pm$ 0.07  & \pmb{69.44} $\pm$ 0.36       & \pmb{3.75} $\pm$ 0.11        & \pmb{70.54} $\pm$ 0.23       \\
				\bottomrule
				\toprule 
				\multirow{3}{*}{Method} & \multicolumn{4}{c}{FLIC} \\
				& \multicolumn{2}{c}{50/100} & \multicolumn{2}{c}{100/200} \\
				& MSE $\downarrow$ & PCK@0.5 (\%) $\uparrow$ & MSE $\downarrow$ & PCK@0.5 (\%)  $\uparrow$\\
				\midrule
				Supervised   & 30.67 $\pm$ 2.07             & 54.17 $\pm$ 2.91             & 23.46 $\pm$ 1.86             & 65.62 $\pm$ 1.53             \\
				\midrule
				DualPose~\cite{xie2021empirical} & 30.99 $\pm$ 0.87 & 48.96 $\pm$ 0.84     & 21.05 $\pm$ 0.62             & 69.79 $\pm$ 0.32             \\
				DualPose+CBE & \underline{27.55} $\pm$ 0.42 & \underline{72.92} $\pm$ 0.93 & \underline{18.44} $\pm$ 0.56 & \underline{86.46} $\pm$ 0.51 \\
				DualPose+DSA & \pmb{25.84} $\pm$ 0.36       & \pmb{78.12} $\pm$ 0.85       & \pmb{16.47} $\pm$ 0.47       & \pmb{87.50} $\pm$ 0.33       \\
				\bottomrule
				\toprule 
				\multirow{3}{*}{Method} & \multicolumn{4}{c}{LSP} \\
				& \multicolumn{2}{c}{100/200} & \multicolumn{2}{c}{200/300} \\
				& MSE $\downarrow$ & PCK@0.5 (\%) $\uparrow$ & MSE $\downarrow$ & PCK@0.5 (\%)  $\uparrow$\\
				\midrule
				Supervised   & 48.38 $\pm$ 1.57             & 16.41 $\pm$ 1.83             & 42.50 $\pm$ 1.06             & 32.81 $\pm$ 1.57             \\
				\midrule
				DualPose~\cite{xie2021empirical} & 44.43 $\pm$ 1.67 & 18.75 $\pm$ 2.31     & 37.64 $\pm$ 1.52             & 32.03 $\pm$ 1.33             \\
				DualPose+CBE & \underline{41.37} $\pm$ 1.08 & \underline{33.59} $\pm$ 1.54 & \underline{34.95} $\pm$ 1.56 & \underline{56.25} $\pm$ 1.05 \\
				DualPose+DSA & \pmb{41.29} $\pm$ 0.95       & \pmb{35.94} $\pm$ 1.33       & \pmb{34.51} $\pm$ 1.32       & \pmb{58.59} $\pm$ 0.98       \\
				\bottomrule
			\end{tabular}%
		}
	\end{table}
	
	The MSE and PCK of our method on the Sniffing, FLIC, and LSP datasets are shown in Table~\ref{tab:Pose_Main_Results}. The comparative results demonstrate that our approach achieves better results than the SOTA methods. For instance, DualPose+DSA achieved a significant improvement over DualPose and DualPose+CBE in PCK@0.2 by 52.77\% and 2.08\% on the Sniffing dataset with 100 data points (including 30 labeled data). Specifically, PCK@0.2 serves as a rigorous metric for evaluating the precision of keypoint localization. Therefore, this significant improvement suggests that the adapters in DSA can effectively reduce correlations between prediction heads, thereby enhancing keypoint localization performance.
	
	\begin{table}[h]
		\caption{Comparison of the performance of SSL methods for various PCK metrics. \label{tab:Pose_PCKs_Results}}
		\centering
		\resizebox{\columnwidth}{!}{%
			\begin{tabular}{@{}lccccc@{}}
				\toprule 
				\multirow{2}{*}{Method} & \multicolumn{5}{c}{Sniffing (60/200)} \\
				& PCK@0.5 (\%) $\uparrow$ & PCK@0.3 (\%) $\uparrow$ & PCK@0.2 (\%) $\uparrow$ & PCK@0.1 (\%) $\uparrow$ & MSE $\downarrow$ \\
				\midrule 
				DualPose~\cite{xie2021empirical} & 79.17 $\pm$ 1.15 & 66.67 $\pm$ 0.98     & 30.56 $\pm$ 1.21             & 0.00 $\pm$ 0.00              & 4.38 $\pm$ 0.07             \\
				DualPose+CBE & \underline{90.28} $\pm$ 0.36 & \underline{84.03} $\pm$ 0.42 & \underline{70.34} $\pm$ 0.35 & \underline{33.33} $\pm$ 1.06 & \underline{4.15} $\pm$ 0.07 \\
				DualPose+DSA & \pmb{93.06} $\pm$ 0.21       & \pmb{86.11} $\pm$ 0.37       & \pmb{70.54} $\pm$ 0.23       & \pmb{35.42} $\pm$ 0.95       & \pmb{3.75} $\pm$ 0.11       \\
				\midrule
				\multirow{1}{*}{Method} & \multicolumn{5}{c}{FLIC (50/100)} \\
				\midrule
				DualPose~\cite{xie2021empirical} & 48.96 $\pm$ 0.84 & 36.46 $\pm$ 1.31     & 17.71 $\pm$ 1.65             & 0.00 $\pm$ 0.00              & 30.99 $\pm$ 0.87            \\
				DualPose+CBE & \underline{72.92} $\pm$ 0.93 & \underline{58.33} $\pm$ 0.82 & \underline{42.71} $\pm$ 1.05 & \underline{28.12} $\pm$ 1.23 & \underline{27.55} $\pm$ 0.42\\
				DualPose+DSA & \pmb{78.12} $\pm$ 0.85       & \pmb{61.46} $\pm$ 0.75       & \pmb{51.04} $\pm$ 0.73       & \pmb{29.17} $\pm$ 0.86       & \pmb{25.84} $\pm$ 0.36      \\
				\midrule
				\multirow{1}{*}{Method} & \multicolumn{5}{c}{LSP (100/200)} \\
				\midrule
				DualPose~\cite{xie2021empirical} & 18.75 $\pm$ 2.31 & 10.16 $\pm$ 2.17     & 4.69 $\pm$ 1.85              & 0.00 $\pm$ 0.00              & 44.43 $\pm$ 1.67            \\
				DualPose+CBE & \underline{33.59} $\pm$ 1.54 & \underline{18.75} $\pm$ 1.51 & \underline{10.94} $\pm$ 0.85 & \underline{4.69} $\pm$ 1.34  & \underline{41.37} $\pm$ 1.08\\
				DualPose+DSA & \pmb{35.94} $\pm$ 1.33       & \pmb{18.75} $\pm$ 1.25       & \pmb{11.72} $\pm$ 0.77       & \pmb{6.25} $\pm$ 1.13        & \pmb{41.29} $\pm$ 0.95      \\
				\bottomrule
			\end{tabular}
		}
	\end{table}
	
	Based on our empirical observations from Table \ref{tab:Pose_PCKs_Results}, DSA demonstrates significant improvements in both the accuracy and stability of keypoint predictions. For instance, on the Sniffing dataset containing 200 data points (including 60 labeled data), DSA led to enhancements of 2.78\%, 2.08\%, 0.2\%, and 2.09\% in PCK@0.5, PCK@0.3, PCK@0.2, and PCK@0.1, respectively, over DualPose+CBE. These improvements with DSA consistently manifest across challenging datasets like FLIC and LSP. Specifically, on the FLIC dataset with 100 data points (including 50 labeled data), DSA resulted in improvements of 5.2\%, 3.13\%, 8.33\%, and 1.05\% in PCK@0.5, PCK@0.3, PCK@0.2, and PCK@0.1, respectively, compared to DualPose+CBE. Similarly, on the LSP dataset, DSA boosted DualPose by 2.35\%, 0.78\%, and 1.56\% in PCK@0.5, PCK@0.2, and PCK@0.1, respectively. Notably, DualPose achieves a 0.0\% score at the strictest PCK@0.1 across all three datasets, whereas DualPose+DSA consistently achieves effective predictions to a certain degree on all three datasets.
	
	In summary, these substantial improvements demonstrate that our method enhances the performance of existing SSL regression by effectively leveraging more accurate pseudo-labels.
	
	\subsection{Ablation}
	\label{ssec:Pose_Estimation_Ablation}
	
	\begin{table}[t!]
		\caption{Results of the ablation experiment in pose estimation with 200 training epochs. \label{tab2_PoseAblationResults}}
		\centering
		\begin{tabular}{@{}lccc@{}}
			\toprule 
			\multirow{2}{*}{Method} & \multicolumn{3}{c}{FLIC (50/100)} \\
			& Adapter    & MSE $\downarrow$             & PCK@0.5 (\%) $\uparrow$      \\
			\midrule
			Supervised   &            & 30.67 $\pm$ 2.07             & 54.17 $\pm$ 2.91             \\
			DualPose     &            & 30.99 $\pm$ 0.87             & 48.96 $\pm$ 0.84             \\
			DualPose+CBE &            & \underline{27.55} $\pm$ 0.42 & \underline{72.92} $\pm$ 0.93 \\
			\midrule
			DualPose+DSA &            & 28.76 $\pm$ 0.65             & 60.37 $\pm$ 1.02             \\
			DualPose+DSA & \checkmark & \pmb{25.84} $\pm$ 0.36       & \pmb{78.12} $\pm$ 0.85       \\
			\bottomrule
			\toprule 
			\multirow{2}{*}{Method} & \multicolumn{3}{c}{LSP (100/200)} \\
			& Adapter    & MSE $\downarrow$             & PCK@0.5 (\%) $\uparrow$      \\
			\midrule
			Supervised   &            & 48.38 $\pm$ 1.57             & 16.41 $\pm$ 1.83             \\
			DualPose     &            & 44.43 $\pm$ 1.67             & 18.75 $\pm$ 2.31             \\
			DualPose+CBE &            & \underline{41.37} $\pm$ 1.08 & \underline{33.59} $\pm$ 1.54 \\
			\midrule
			DualPose+DSA &            & 43.21 $\pm$ 1.15             & 20.23 $\pm$ 1.73             \\
			DualPose+DSA & \checkmark & \pmb{41.29} $\pm$ 0.95       & \pmb{35.94} $\pm$ 1.33       \\
			\bottomrule
			\toprule 
		\end{tabular}
	\end{table} 
	
	The ablation study results of our method on SSL pose estimation tasks are presented in Table~\ref{tab2_PoseAblationResults}.
	
	Without adapters, due to the absence of the decorrelation loss function, DualPose+DSA decreased by 12.55\% compared to DualPose+CBE on the FLIC dataset with 100 data points (including 50 labeled data). Moreover, on the LSP dataset with 200 data points (including 100 labeled data), the decrease was 13.36%.
	
	With adapters, on the FLIC dataset with 100 data points (including 50 labeled data), DualPose+DSA performed 17.75\% better than without adapters, outperforming DualPose+CBE by 5.2\%. Furthermore, on the LSP dataset with 200 data points (including 100 labeled data), DualPose+DSA performed 15.71\% better than without adapters, surpassing DualPose+CBE by 2.35\%.
	
	These experiments indicate that adapters successfully substitute for the decorrelation loss function, effectively reducing the correlation between different prediction heads and consolidating multiple pseudo-labels into more accurate ones in SSL regression tasks.
	
	\subsection{Computation Cost Analysis}
	\label{ssec:Pose_Estimation_Computation_Cost_Analysis}
	
	\begin{table}[h]
		\caption{The computational costs of DualPose+CHE and DualPose+DSA were compared on the FLIC dataset (50/100). The ATTPE was measured on a GPU 4090. \label{tab:Pose_Estimation_Cost}}
		\centering
		\resizebox{\columnwidth}{!}{%
			\begin{tabular}{@{}lccccc@{}}
				\toprule 
				Method & Model Parameters $\downarrow$ & FLOPs $\downarrow$ & MACs (Millions) $\downarrow$ & ATTPE (Seconds) $\downarrow$ & PCK@0.5 (\%) $\uparrow$ \\
				\midrule
				DualPose+CBE & 9.72M & 11.661G & 2441424.66 & 28.82 & 72.92      \\ 
				DualPose+DSA & 9.72M & 11.661G & 2441424.66 & \pmb{28.13} & \pmb{78.12}\\ 
				\bottomrule
			\end{tabular}%
		}
	\end{table}
	
	Adapters consist of several $1\times1$ convolutional layers and activation functions, with a relatively small number of parameters, thus not significantly increasing the computational cost. Table~\ref{tab:Pose_Estimation_Cost} illustrates this point. Compared to CBE, DSA is nearly equivalent in terms of model parameters, Floating Point Operations (FLOPs), Multiply-Accumulate Operations (MACs), etc., while achieving a 5.2\% improvement in performance on the FLIC dataset (100 data points with 50 labels).
	
	Furthermore, by avoiding the complex computation process of the decorrelation loss function, DSA exhibits a reduced training time of 0.69 seconds per epoch on a GPU 4090 compared to CBE.
	
	\section{Experiments on SSL Classification}
	\label{sec:Experiments_on_Classification}
	
	\begin{table*}[t]
		\caption{The error rates (\%) of the SSL classification on the CIFAR-10/100 datasets with 200 training epochs are presented. ``X+CBE'' denotes the CBE network, and ``X+DSA'' denotes our DSA network. \label{tab:Comparisons}}
		\centering
		\begin{tabular}{@{}lcccccc@{}}
			\toprule
			\multirow{2}{*}{Method} & \multicolumn{3}{c}{CIFAR-10}  & \multicolumn{3}{c}{CIFAR-100}\\
			& 40 $\downarrow$ & 250 $\downarrow$ & 4000 $\downarrow$ & 400 $\downarrow$ & 2500 $\downarrow$ & 10000 $\downarrow$ \\
			\midrule
			FreeMatch~\cite{wang2022freematch} & 14.85 $\pm$ 0.51 & 5.85 $\pm$ 0.23 & 4.95 $\pm$ 0.05 & 44.41 $\pm$ 0.74 & 28.04 $\pm$ 0.28 & 22.37 $\pm$ 0.09 \\
			FreeMatch+CBE & \underline{11.45} $\pm$ 0.28 & \pmb{5.25} $\pm$ 0.20 & \pmb{4.55} $\pm$ 0.05 & \underline{43.64} $\pm$ 0.62 & \pmb{26.85} $\pm$ 0.25 & \underline{22.33} $\pm$ 0.10 \\
			FreeMatch+DSA & \pmb{6.10} $\pm$ 0.26 & \underline{5.80} $\pm$ 0.23 & \underline{4.83} $\pm$ 0.15 & \pmb{42.93} $\pm$ 0.33 & \underline{27.18} $\pm$ 0.23 & \pmb{22.19} $\pm$ 0.08 \\
			\bottomrule
			\toprule
		\end{tabular}
	\end{table*}
	
	We evaluate the efficacy of our approach on the CIFAR-10/100 datasets to demonstrate the improvement achieved by combining classical SSL methods with our DSA. Specifically, FreeMatch is an improvement over the classic weak-strong method, FixMatch, and represents the current state-of-the-art (SOTA) in SSL technology. Note that when an SSL method is combined with our approach, it requires only one modification: using the DSA class library provided by us to modify the classification model into the DSA style.
	
	\subsection{Configurations}
	\label{ssec:Classification_Configurations}
	
	The classification model used in all experiments was Wide ResNet with a widen-factor of 2 for the CIFAR-10 dataset and 6 for the CIFAR-100 dataset, respectively.
	
	For a fair comparison, the same hyperparameters as in FreeMatch were used for all experiments. Specifically, all experiments were performed using a standard Stochastic Gradient Descent (SGD) optimizer with a momentum of 0.9~\cite{sutskever2013importance, polyak1964some} and Nesterov momentum enabled~\cite{dozat2016incorporating}. The learning rate was set to 0.03. The batch size for labeled data was 32, and the batch size for unlabeled data was 224. Data augmentation on labeled data included random horizontal flipping and random cropping. For unlabeled data, weak data augmentation was similar, but strong data augmentation included RandAugment~\cite{cubuk2020randaugment}.
	
	The confidence threshold of FreeMatch was dynamically controlled by the Self-Adaptive Threshold (SAT) technique. The number of multi-heads in MHE, CBE and our DSA was 5.
	
	\subsection{Comparisons}
	\label{ssec:Classification_Comparisons}
	
	Table~\ref{tab:Comparisons} shows the performances of SSL methods. Compared to existing state-of-the-art (SOTA) methods, DSA demonstrates superior performance under limited labeled data conditions. Specifically, on the CIFAR-10 dataset (with 40 labels), FreeMatch+DSA improves upon FreeMatch and FreeMatch+CBE by 8.75\% and 5.35\%, respectively. Similarly, on the CIFAR-100 dataset (with 400 labels), FreeMatch+DSA also surpasses FreeMatch and FreeMatch+CBE by 1.48\% and 0.71\%, respectively, demonstrating the effectiveness of DSA in limited labeled data scenarios.
	
	The CBE method employs the LB loss to maintain low correlation between multiple prediction heads, but this LB loss is detrimental to the feature representation of the model. However, our DSA method directly reduces correlation through a structure-based decorrelation manner, which is more conducive to feature representation learning.
	
	\subsection{Ablation}
	\label{ssec:Classification_Ablation}
	
	\begin{table}[h]
		\caption{Ablation experiments were conducted on CIFAR-10@40 with 200 training epochs.\label{tab:Ablation}}
		\centering
		\begin{tabular}{@{}lcc@{}}
			\toprule 
			Method        & Adapter              & Error rate(\%) $\downarrow$  \\
			\midrule
			FreeMatch~\cite{wang2022freematch} & & 14.85 $\pm$ 0.51             \\
			FreeMatch+CBE &                      & \underline{11.45} $\pm$ 0.28 \\
			\midrule
			FreeMatch+DSA &                      & 13.35 $\pm$ 0.42             \\
			FreeMatch+DSA & \checkmark           &\pmb{6.10} $\pm$ 0.26         \\
			\bottomrule
			\toprule
		\end{tabular}
	\end{table}
	
	The ablation study results of our method on SSL classification tasks are presented in Table~\ref{tab:Ablation}.
	
	Without adapters, due to the absence of the decorrelation loss function, FreeMatch+DSA decreased by 1.9\% compared to FreeMatch+CBE on the CIFAR-10 dataset (with 40 labels).
	
	With adapters, on the CIFAR-10 dataset (with 40 labels), FreeMatch+DSA performed 7.25\% better than without adapters, resulting in a 5.35\% improvement over FreeMatch+CBE.
	
	As depicted in Table~\ref{tab:Ablation}, the adapters in DSA successfully substitute for the decorrelation loss function, effectively reducing the correlation between different prediction heads. This characteristic is crucial for achieving ensemble gains and consolidating multiple pseudo-labels into more accurate ones in SSL tasks.
	
	\section{Experiments on Supervised Classification}
	\label{sec:Experiments_on_Classification_sup}
	
	We evaluate the efficacy of our approach on two noisy datasets, CIFAR-10C~\cite{hendrycks2019benchmarking} and Animal-10N~\cite{song2019selfie}.
	
	\subsection{Dataset}
	\label{ssec:Classification_Dataset_sup} 
	
	\textbf{CIFAR-10C.} The CIFAR-10C~\cite{hendrycks2019benchmarking} dataset applies 15 common image corruptions, such as Gaussian noise, impulse noise, motion blur, frost, etc., to the CIFAR-10 test set. Each type of corruption is characterized by five severity levels, as these corruptions can occur at different intensities.
	
	\textbf{Animal-10N.} Animal-10N~\cite{song2019selfie} is a real-world dataset where noisy labels are present in the training set. It serves as a benchmark that contains 10 animal classes with confusing appearances. The training set size is 50,000, and the test set size is 5,000. The estimated label noise ratio of the training set is 8\%.
	
	\subsection{Configurations}
	\label{ssec:Classification_Configurations_sup}
	
	The classification model used in all experiments was Wide ResNet with a widen-factor of 2 for the CIFAR-10C and Animal-10N datasets.
	
	All experiments were performed using a standard Stochastic Gradient Descent (SGD) optimizer with momentum 0.9~\cite{sutskever2013importance, polyak1964some} and Nesterov momentum~\cite{dozat2016incorporating} enabled. The learning rate was set to 0.03. The batch size for the labeled data was 64. Data augmentation on the labeled data included random horizontal flipping and random cropping.
	
	The number of multi-heads in MHE, CBE, and our DSA was 5.
	
	In the context of supervised learning, it is noteworthy that DSA does not utilize the pseudo-label generation and supervision process during the training phase, but rather relies solely on the ensemble prediction framework in the inference phase. Therefore, the experimental design for supervised learning is specifically aimed at evaluating the performance of DSA's ensemble prediction framework in the inference phase.
	
	\subsection{Comparisons}
	\label{ssec:Classification_Comparisons_sup}
	
	\begin{table}[h]
		\caption{The error rates (\%) of the supervised classification on the CIFAR-10C and Animal-10N datasets, obtained after 400 training epochs, are presented. \label{tab:Comparisons_sup}}
		\centering
		\begin{tabular}{@{}lcc@{}}
			\toprule
			Method  & CIFAR-10C $\downarrow$ & Animal-10N $\downarrow$ \\
			\midrule
			VGG-11     & 11.72            & 22.36             \\
			VGG-13     & 9.40             & 19.38             \\
			VGG-16     & 9.14             & 19.04             \\
			VGG-19     & 9.40             & 19.32             \\
			ResNeXt    & 10.19            & 18.62             \\
			WideResNet & 7.81             & 17.18             \\
			\midrule
			MHE        & \underline{7.87} & 17.14             \\
			CBE        & \pmb{7.77}       & \underline{15.78} \\
			DSA        & 8.33             & \pmb{15.70}       \\
			\bottomrule
			\toprule
		\end{tabular}
	\end{table}
	
	As shown in Table \ref{tab:Comparisons_sup}, in supervised classification tasks on the CIFAR-10C dataset with image noise, DSA exhibits inferior performance compared to MHE, with a classification error rate higher by 0.46\%. This phenomenon can be attributed to the fact that, within the supervised learning framework, DSA does not employ a pseudo-label generation mechanism during the training phase, thus failing to effectively suppress image noise interference during training. DSA relies solely on the ensemble prediction structure in the inference phase, which is susceptible to noise in the images, adversely affecting the accuracy of ensemble predictions.
	
	However, on the Animal-10N dataset with label noise, DSA demonstrates a clear advantage. Compared to MHE, DSA achieves a significant reduction in classification error rate by 1.44\% (see Table \ref{tab:Comparisons_sup}). This result fully indicates that the ensemble prediction structure adopted by DSA in the inference phase can effectively mitigate the negative effects of label noise, thereby enabling more precise predictions.
	
	\subsection{Ablation}
	\label{ssec:Classification_Ablation_sup}
	
	\begin{table}[h]
		\caption{The ablation results are presented. \label{tab:Classification_Ablation_sup}}
		\centering
		\begin{tabular}{@{}lcc@{}}
			\toprule
			Method     & Adapter    & Animal-10N $\downarrow$ \\
			\midrule
			WideResNet &            & 17.18             \\
			\midrule
			MHE        &            & 17.14             \\
			CBE        &            & \underline{15.78} \\
			\midrule
			DSA        &            & 17.03 \\
			DSA        & \checkmark & \pmb{15.70}       \\
			\bottomrule
			\toprule
		\end{tabular}
	\end{table}
	
	The ablation study results of our method on supervised classification tasks are presented in Table~\ref{tab:Classification_Ablation_sup}.
	
	Without adapters, DSA showed only a 0.11\% increase compared to CBE on the Animal-10N dataset, due to the absence of the decorrelation loss function. With adapters, DSA performed 1.33\% better than without adapters and outperformed FreeMatch+CBE by 1.44\%.
	
	These findings indicate that employing an ensemble method with multiple prediction heads can effectively mitigate the adverse effects of label noise. Meanwhile, adapters significantly reduce the correlation among individual predictors within the ensemble framework, further enhancing this mitigation process.
	
	\section{Analysis}
	\label{sec:Analysis}
	
	\subsection{The Decorrelation Effect of Adapters in DSA}
	\label{ssec:Decorrelation_Effect_Adapters}
	
	\begin{figure*}[t]
		\centering
		\subfloat{\includegraphics[width=1.3in]{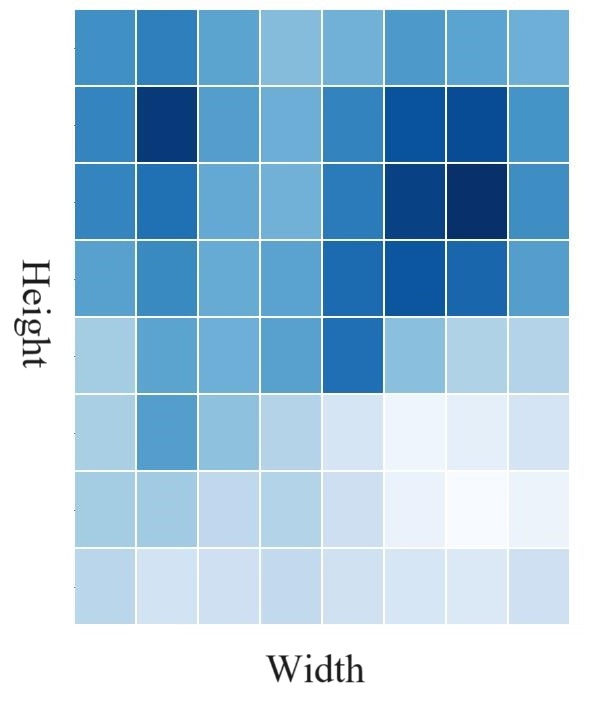}}
		\subfloat{\includegraphics[width=1.3in]{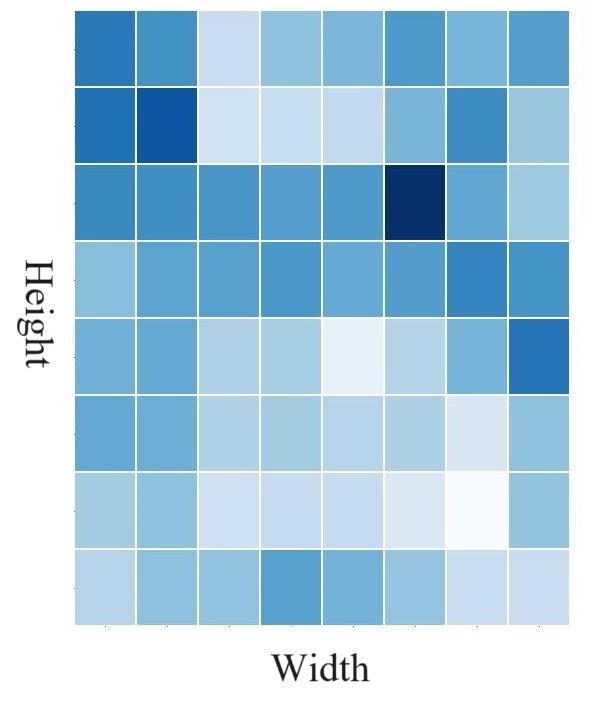}}
		\subfloat{\includegraphics[width=1.3in]{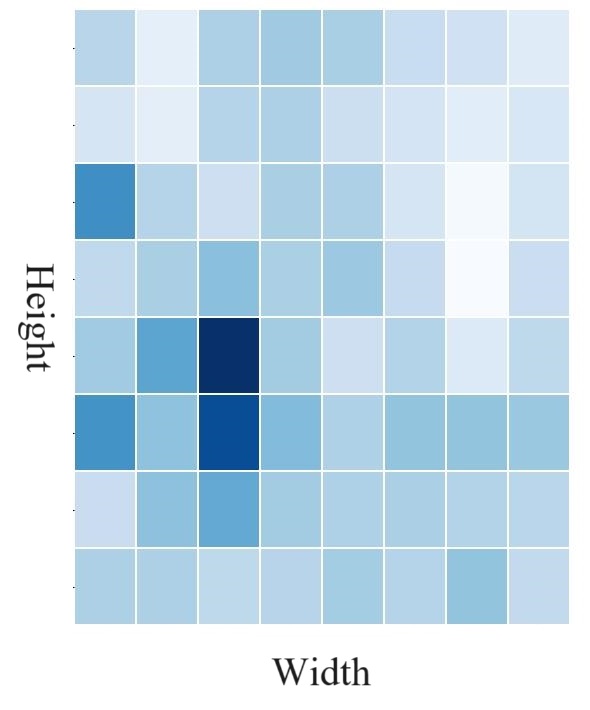}}
		\subfloat{\includegraphics[width=1.3in]{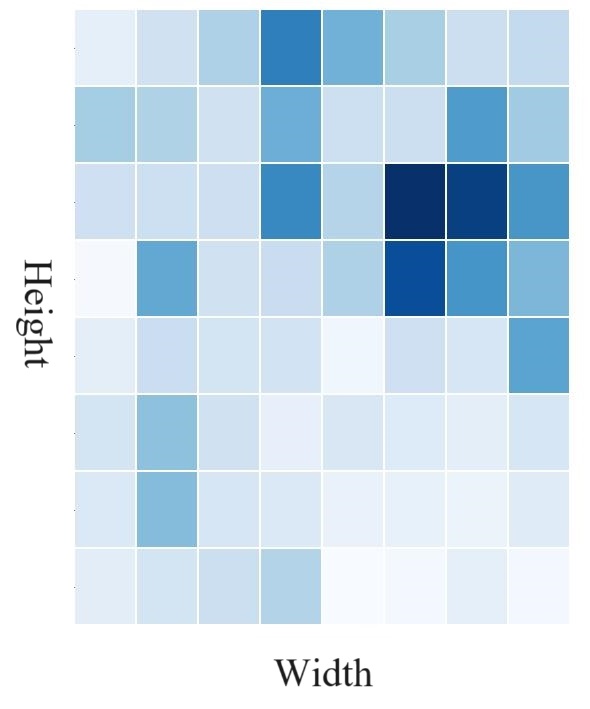}}
		\subfloat{\includegraphics[width=1.3in]{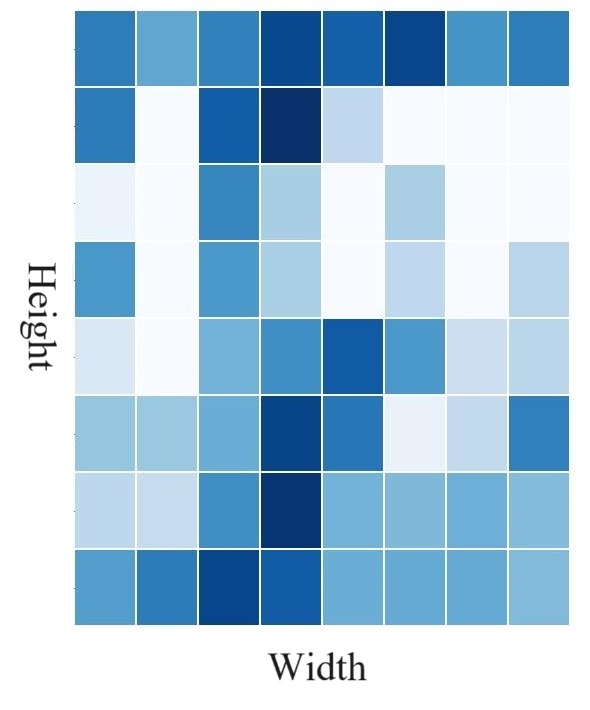}}
		\caption{The comparison results of feature maps output by the $5$ adapters in FreeMatch+DSA on the same channel are presented. The displayed features are 0.25x scaled versions of the original features. \label{fig:DSA_featuremap}}
	\end{figure*}
	
	\begin{figure}[h]
		\centering
		\includegraphics[width=2.5in]{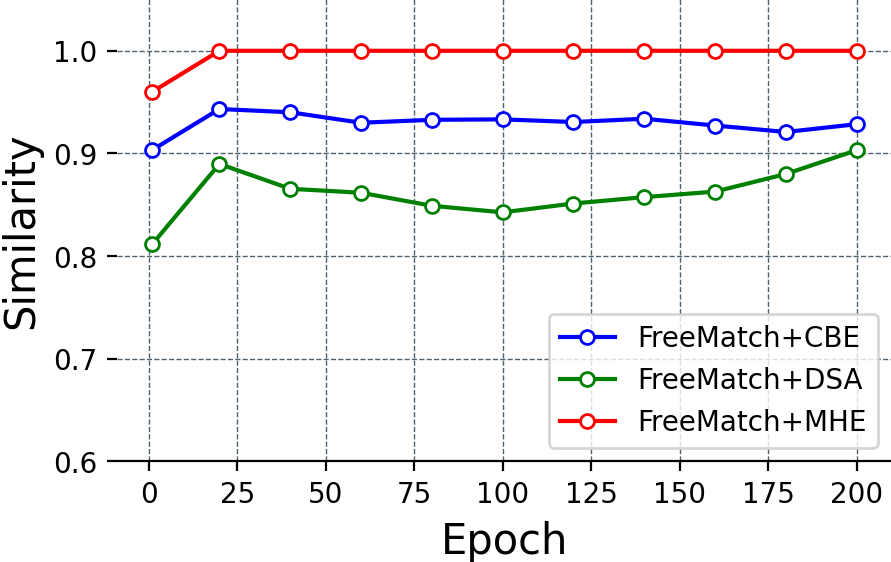}
		\caption{
			Comparing the similarity among the prediction heads of FreeMatch+MHE, FreeMatch+CBE, and FreeMatch+DSA. 
			\label{fig:DSA_prediction_similarity}}
	\end{figure}
	
	In the DSA framework, we employ a set of structurally diverse adapters to implement a structured decorrelation process that does not rely on a loss function. The core of this process is to utilize these adapters to map the same input features into different feature spaces, thereby effectively reducing the correlation between the prediction results of different prediction heads. To validate the effectiveness of our method in decorrelation, we conduct an analysis from two dimensions: the feature representations output by each adapter and the prediction similarity among different prediction heads.
	
	The feature maps output by each adapter are shown in Fig.~\ref{fig:DSA_featuremap}. As observed, the feature representation of the same sample undergoes different mapping transformations through various adapters, resulting in distinct feature representations. This process enables different prediction heads in DSA to predict the same sample from diverse dimensions, effectively reducing the correlation between the prediction heads. 
	
	To further analyze the effectiveness of the adapter mechanism, we calculated the average prediction similarity between the prediction heads, as shown in Fig.~\ref{fig:DSA_prediction_similarity}. Unlike FreeMatch+MHE, which tends to produce identical results, both FreeMatch+CBE and FreeMatch+DSA exhibit the ability to generate diverse prediction results by reducing the correlation among different prediction heads. Notably, FreeMatch+DSA demonstrates a more pronounced advantage in decorrelation. Specifically, as shown in Fig.~\ref{fig:DSA_prediction_similarity}, throughout the entire training process, the similarity between the prediction heads of FreeMatch+DSA remains at a lower level compared to FreeMatch+CBE. This observation further indicates that the adapter mechanism in DSA effectively reduces the correlation among prediction heads through a structured approach.
	
	\section{Conclusions and Future Work}
	\label{sec:Conclusions_and_Future_Work}
	
	In this paper, we introduce a lightweight, loss-function-free, and architecture-agnostic ensemble learning approach, termed Decorrelating Structure via Adapters (DSA). DSA leverages a structure-based decorrelation method to reduce the correlation between prediction heads, thereby achieving ensemble gains. As it does not require additional loss functions, DSA can avoid complex hyperparameter tuning processes, over-decorrelation, conflicts in loss function optimization, and significant increases in computational costs. Our method unifies both classification and regression tasks, achieving better performance on classification datasets such as CIFAR-10/100, as well as pose estimation datasets including Sniffing, FLIC, and LSP. Furthermore, on noisy datasets like CIFAR-10C and Animal-10N, our method also demonstrates better performance, highlighting the role of low-bias and low-variance predictions in mitigating the impact of noisy labels.
	
	In our future work, we plan to address the following objectives: 1) In SSL, confidence scores are not suitable for assessing the accuracy of pseudo-labels. Therefore, we aim to develop a new method for evaluating pseudo-labels more reasonably, avoiding reliance on confidence scores. 2) The training of pseudo-label (PL) methods is inefficient. We plan to leverage the dynamic nature of the training process to filter a subset of informative unlabeled samples at each training epoch, thereby improving training efficiency.
	
	\bibliography{reference}
	\bibliographystyle{IEEEtran}

\end{document}